\documentclass[11pt]{article}

\usepackage[final]{acl}
\usepackage{times}
\usepackage{latexsym}
\usepackage[T1]{fontenc}
\usepackage[utf8]{inputenc}
\usepackage{microtype}
\usepackage{inconsolata}
\usepackage{amsmath}
\usepackage{array}
\usepackage{booktabs}
\usepackage{float}
\usepackage{stfloats}
\usepackage{tikz}
\usetikzlibrary{positioning,arrows.meta,calc}
\usepackage{pgfplots}
\pgfplotsset{compat=1.18}

\tikzset{
  llmbox/.style={draw, dashed, rounded corners=3pt, fill=black!6,
    align=center, font=\footnotesize, inner sep=4.5pt, minimum height=8.5mm},
  detbox/.style={draw, fill=black!14, align=center, font=\footnotesize,
    inner sep=4.5pt, minimum height=8.5mm},
  artifact/.style={font=\footnotesize\itshape},
  flow/.style={-{Stealth[length=2.2mm]}, semithick},
  fignote/.style={font=\scriptsize, align=center, text=black!70},
}

\newcommand{\system}{\textnormal{\textsc{RevGuard}}}
\newcommand{\verified}{\textsc{Verified}}
\newcommand{\repair}{\textsc{Verified-G}}
\newcommand{\compiler}{\textsc{Compiler}}
\newcommand{\reference}{\textsc{Reference}}

\title{\system: Transactional, Version-Bound Contracts for LLM
Document Editing}
\author{
  Jinkun Chen$^{1}$ \quad Xuemin Yu$^{1}$ \quad
  Sijia Han$^{2}$ \quad Vlado Keselj$^{1}$ \\
  $^{1}$Faculty of Computer Science, Dalhousie University \\
  $^{2}$Meta \\
  \texttt{\{jinkun.chen,xuemin.yu\}@dal.ca} \\
  \texttt{hansijia@meta.com} \quad \texttt{vlado@cs.dal.ca}
}
\hypersetup{
  pdftitle={RevGuard: Transactional, Version-Bound Contracts for LLM Document Editing},
  pdfauthor={Jinkun Chen, Xuemin Yu, Sijia Han, Vlado Keselj}
}

\begin{document}
\maketitle

\begin{abstract}
Large language model (LLM) agents increasingly modify persistent shared
artifacts, but natural-language instructions do not define safe write
authority when a workspace changes between read and commit.  We present
\system, a transactional layer that compiles instructions into immutable,
version-bound edit contracts and commits only the exact previewed state.
Across 144 episodes on licensed, compilable \LaTeX{} repositories, every
episode terminated and no injected fault committed.  An LLM compiler proposes
contracts, an independent LLM verifier reviews them, and deterministic
projection installs least privilege.  Evaluation separates setup reliability,
commit safety, and task utility.  Verification improved normalized
authorization in 7/48 paired contracts without observed protection loss, but
did not establish a task-utility benefit: verified and compiler-only contracts
succeeded in 46/48 episodes, and strict non-inferiority remained unresolved.
In a descriptive 900-cell cross-model stress test, contract protocols committed
zero faults, unmanaged patching baselines committed over one hundred each, and
safe whole-file replacement succeeded in under five percent of tasks.  One
fail-closed setup left the stress test inconclusive; a versioned repair passed
targeted compatibility tests.  A Git three-way baseline produced 176 clean
merges across 180 episodes, yet committed 123 controlled faults.  These results
support containment of tested faults in localized \LaTeX{} edits, not semantic
or factual correctness.
\end{abstract}

\section{Introduction}

Tool-using language models increasingly act through external interfaces rather
than only generate text~\citep{yao2023react,schick2023toolformer}.  Software agents
also modify persistent repositories~\citep{yang2024sweagent,jimenez2023swebench};
stateful tool benchmarks expose dependencies between actions~\citep{lu2024toolsandbox},
and delegated-work studies find that document fidelity can degrade over long
editing workflows~\citep{laban2026delegate}.  These settings share a systems
problem: an agent's natural-language objective does not itself define the
write authority under which a persistent state transition should commit.

The problem is sharper in collaborative editing.  Consider an agent asked to
change a paper's title.  While it works, a co-author revises a section of the
same file; the agent writes back the full contents it read earlier, and the
revision vanishes from a commit that compiles and satisfies the requested
change.  Nothing in the instruction, the patch, or the compiler output marks
this lost update as a failure.  A staged patch can also change an unrelated
file, delete a citation, or relax its own guard.  Prompting the model to
``be careful'' leaves the same model
responsible for proposing the change, interpreting its authority, and
deciding whether the result is safe.

We study a different abstraction: treat an agentic edit as a
transaction~\citep{gray1981transaction} authorized by an immutable,
version-bound contract.  A contract names a base
version, read and write sets, allowed files, a change budget, and exact
fragments that must survive.  The agent may inspect, stage, preview, refresh,
rollback, and commit, but cannot silently widen the installed contract while
content is staged.  Deterministic checks mediate the final state transition.

The difficult part is not enforcing a contract once written; it is compiling
a useful least-privilege contract from an ambiguous instruction.  A contract
that is too broad admits collateral changes, while one that is too narrow
causes safe but unproductive refusals.  \system{} uses an LLM compiler to
propose a candidate, a separately prompted LLM verifier to review its semantics,
and a deterministic projection to produce the final contract.  The projection
is fail-closed for measured protections but removes redundant read-only
constraints and exact replacement conflicts.

This separation leads to the organizing principle of our evaluation.  A
system may prevent an unsafe commit yet fail the requested edit; it may also
have a sound transaction layer but fail to install a usable contract.  We
therefore report \emph{setup reliability}, \emph{commit safety}, and
\emph{task utility} independently.  Contract protection and authorization
provide paired mechanism endpoints between setup and execution.

Least privilege, proposal--review separation, and optimistic concurrency are
established principles, not our individual novelty.  We instead contribute a
mutation-specific interface: an edit contract binds instruction-conditioned
write authority and preservation invariants to a base version and exact
previewed state.  Independent model review plus deterministic projection
exposes paired protection and authorization endpoints rather than treating
successful execution as evidence of safe authority.  Finally, our
reproducible API-only evaluation on compilable document repositories reports
setup reliability, commit safety, task utility, protection, and authorization
separately.

The evaluation stages serve distinct roles.  The primary test supports fault
containment and authorization refinement but leaves strict task-success
non-inferiority unresolved.  The cross-model stress test adds unmanaged
baselines: contracts committed zero controlled faults, coarse patches more
than one hundred per package, and safe whole-file replacement rarely
succeeded.  One fail-closed setup keeps its all-cell claim inconclusive; a
versioned repair provides only targeted compatibility evidence.  A separate
post-review Git arm isolates the merge boundary: all textual-concurrency cells
succeeded, yet clean merges elsewhere admitted controlled authorization and
preservation faults.

\section{Problem and Threat Model}

\subsection{Collaborative edit transaction}

Let $S_v$ be a multi-file workspace at version $v$, and let $I$ be a user's
natural-language edit instruction.  An editor agent produces a staged patch
$\Delta$ after a sequence of observations and tool calls.  Concurrent events
$E$ may change the workspace after the agent first reads it.  A successful
system must decide whether applying $\Delta$ to the current state is both
\emph{authorized}, meaning the patch is within the scope intended by $I$, and
\emph{valid}, meaning the resulting artifact satisfies structural and
task-specific requirements.

These properties are distinct.  Compilation is a validity signal, not a
complete authorization decision.  Conversely, a safe refusal can preserve
authority while failing the requested task.

\subsection{Adversary and failures}

We do not assume a malicious model.  We consider accidental failures arising
from stochastic generation, stale observations, coarse file-writing tools,
incorrect escaping, over-broad or over-constrained contracts, and concurrent
changes.  Controlled benchmark events can alter staged content or the
underlying workspace between editing and commit, a time-of-check-to-time-of-use
hazard~\citep{bishop1996toctou}.  The primary safety event is
an injected fault that survives into the committed workspace.

The local transaction layer, evaluator, and benchmark scheduler are trusted.
The editor, contract compiler, and semantic verifier are untrusted probabilistic
components.  Model providers are accessed through an API gateway; the system
does not require local model weights or accelerator hardware.

\subsection{Desired properties}

Our safety requirements combine version binding, write isolation, and
preservation.  A contract applies to one observed base state, so version drift
requires refresh or abort.  Only authorized paths and a bounded patch may
commit, and explicit document fragments and cross-file dependencies must
survive the transaction.  We evaluate these guarantees alongside totality,
which requires contract projection and every episode to terminate under a
hard watchdog, and utility, which asks whether the resulting restrictions
still permit the requested edit.

\section{\system{} Design}

\begin{figure*}[t]
\centering
\begin{tikzpicture}
  \node[font=\footnotesize\itshape, align=center] (inst)
    {instruction $I$,\\ bounded view of $S_v$};
  \node[llmbox, right=6mm of inst] (comp) {contract\\ compiler};
  \node[llmbox, right=11mm of comp] (ver) {contract\\ verifier};
  \node[detbox, right=11mm of ver] (proj) {projection $\Pi$\\ + grounding};
  \node[detbox, right=7mm of proj] (install) {install $C_f$\\ (version-bound)};

  \draw[flow] (inst) -- (comp);
  \draw[flow] (comp) -- node[above, artifact] {$C_c$} (ver);
  \draw[flow] (ver) -- node[above, artifact] {$C_v$} (proj);
  \draw[flow] (proj) -- (install);

  \node[fignote, above=1.4mm of comp] {schema + scope checks;\\ one repair turn};
  \node[fignote, above=1.4mm of proj] {two invalid attempts $\Rightarrow$\\ fail-closed abstention (\repair{})};

  \node[llmbox] (editor) at ($(inst.center)+(0,-27mm)$) {editor};
  \node[detbox, right=12mm of editor] (ws) {versioned\\ workspace $S_v$};
  \node[detbox, right=11mm of ws] (preview) {preview:\\ guards $A,B,P$};
  \node[detbox, right=12mm of preview] (commit)
    {commit exactly the\\ previewed txn;\\ drift $\Rightarrow$ refuse};
  \node[detbox, below=4.5mm of commit] (rollback) {rollback};

  \draw[{Stealth[length=2.2mm]}-{Stealth[length=2.2mm]}, semithick]
    (editor) -- node[below=0.5mm, fignote] {read,\\ stage} (ws);
  \draw[flow] (ws) -- node[above, artifact] {$\Delta$} (preview);
  \draw[flow] (preview) -- node[above, artifact] {txn id} (commit);
  \draw[flow] (preview.south) to[out=-75, in=180]
    node[pos=0.35, below left=-0.5mm, fignote] {or} (rollback.west);

  \draw[flow] (install.south) -- ++(0,-6.5mm) -|
    node[pos=0.18, below, fignote] {enforced at preview and commit} (preview.north);

  \node[fignote, below=1.4mm of ws] {concurrent events $E$\\ $\Rightarrow$ version drift};

  \node[llmbox, minimum height=5mm, font=\scriptsize, inner sep=2.5pt, anchor=west]
    (leg1) at ($(editor.west)+(14mm,-24mm)$) {LLM};
  \node[fignote, right=1.5mm of leg1, anchor=west] (leg1t) {probabilistic, untrusted};
  \node[detbox, minimum height=5mm, font=\scriptsize, inner sep=2.5pt,
    right=12mm of leg1t] (leg2) {deterministic};
  \node[fignote, right=1.5mm of leg2, anchor=west] {trusted transaction layer};
\end{tikzpicture}
\caption{\system{} separates probabilistic proposal from deterministic
enforcement.  An LLM compiler and an independently prompted LLM verifier
propose a contract for instruction $I$; a deterministic projection grounds
and installs the final contract $C_f$.  The editor stages a minimal edit against the versioned workspace, and
deterministic preview and commit enforce the write scope, change budget, and
preservation atoms under concurrent events.}
\label{fig:architecture}
\end{figure*}
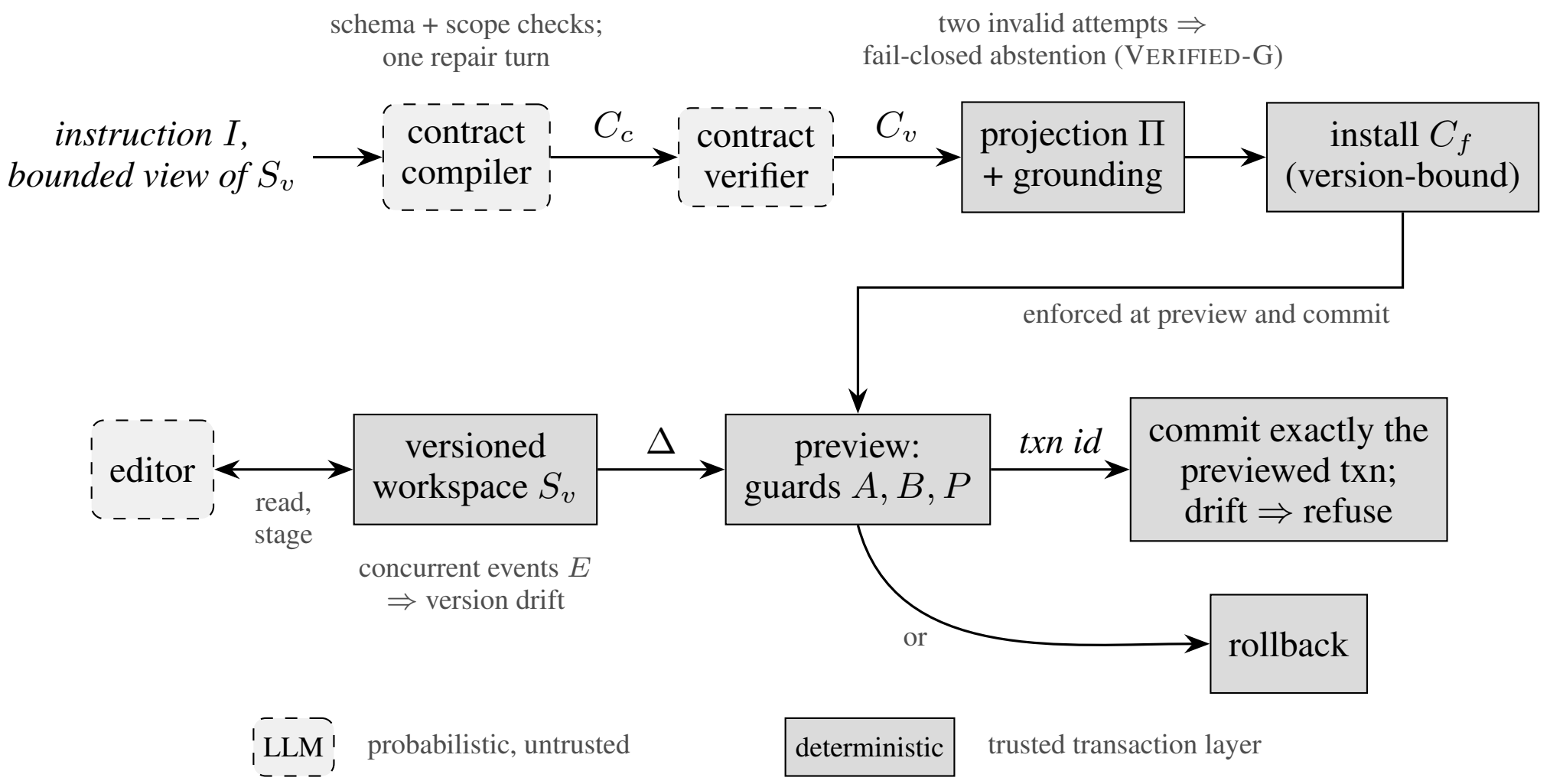

\subsection{Edit contracts}

An edit contract is a tuple
\[
C=(v,R,W,A,B,P),
\]
where $v$ is the base version, $R$ and $W$ are read and write sets, $A$ is the
allowed-file set, $B$ is a maximum changed-line budget, and $P$ is a set of
path-bound exact preservation atoms.  The transaction layer validates that
$W\subseteq A$, that named paths exist, that preservation atoms are grounded
in the authorized base state, and that the staged diff respects $A$, $B$, and
$P$.  Figure~\ref{fig:contract-example} instantiates the enforced components
for one benchmark task.

\begin{figure}[t]
\centering
\fbox{\begin{minipage}{0.93\linewidth}\footnotesize
\textbf{Instruction $I$.} ``Change the \LaTeX{} title from ``A Baseline
Paper'' to exactly ``A Transaction-Safe Paper''.  Preserve the author and all
body content.  Make the smallest possible edit and ensure the project still
compiles.''

\medskip
\textbf{Reference contract.}
$v$: the observed base version;\quad
$W=A=\{\texttt{main.tex}\}$;\quad
$B$: at most 2 changed lines.  The preservation set $P$ contains
\texttt{\detokenize{\author{Ada Researcher}}},
\texttt{\detokenize{\label{sec:introduction}}}, and
\texttt{\detokenize{\label{sec:method}}}.
\end{minipage}}
\caption{An example benchmark instruction with its task-author reference
contract, showing the components enforced at preview and commit: a two-line
change budget over a single authorized file and three exact preservation
atoms, bound to the observed base version.}
\label{fig:contract-example}
\end{figure}

The contract is external to the editor.  Managed protocols do not expose
benchmark assertions, injected-fault metadata, or the task-author reference
contract to the compiler or verifier.  Once content is staged, the editor
cannot redeclare a weaker contract.

\subsection{Transaction lifecycle}

Figure~\ref{fig:architecture} shows the episode lifecycle: the agent reads a
versioned workspace, installs a managed contract, stages a minimal edit,
optionally receives a concurrent event, previews deterministic guards, and
then either commits exactly the previewed transaction or rolls back.

The preview produces a transaction identifier derived from the base state,
contract, and staged content.  Commit requires the same identifier and fails
if the workspace has drifted.  An injected staged fault must be repaired and
re-previewed; it cannot ride on an earlier successful preview.

\subsection{Compiler, verifier, and projection}

Throughout the paper, \system{} names the transaction architecture rather
than one frozen model package.  \verified{} is the frozen package used in the
primary and cross-model studies; \repair{} is its separately versioned grounding
repair.

The \compiler{} protocol asks one LLM to compile $I$ and a bounded view of the
workspace into $C_c$.  Schema validation, grounding, scope checks, and a
minimality linter reject malformed output, with one explicit repair turn.

The \verified{} protocol sends $I$, the bounded workspace, and $C_c$ to an
independently prompted verifier.  This proposal--review separation builds on
verifier ranking and LLM-as-judge evaluation~\citep{cobbe2021verifiers,zheng2023judging},
rather than same-model self-refinement~\citep{madaan2023selfrefine}.  The verifier
returns a proposed $C_v$.  A deterministic
projection $\Pi(C_c,C_v,I,S_v)$ then produces the installed contract $C_f$.
The projection normalizes write and allowed-file sets, atomizes preservation
clauses and removes redundant read-only atoms, derives exact replacement
budgets where possible, and retains candidate protection on writable files
unless it conflicts with the explicitly authorized old side of a replacement.
Grounding is validated only after this normalization.

Projection is implemented as a total function over adversarial strings.  A
process watchdog provides an independent episode-level termination boundary.

The registered all-cell stress test later exposed a representation boundary
in the original package: a verifier twice returned an instruction-grounded but
base-ungrounded overescaped preservation atom.  We preserve that method and
result and define \repair{}.  For verifier-only preservation, its
projection retains an atom only if the trimmed string occurs exactly in the
corresponding base file; protections from the already validated compiler
candidate are then restored.  If two verifier attempts still cannot produce a
valid contract, the runner persists an explicit fail-closed abstention, calls
no editor, installs no contract, and scores authorization and task success as
false.  The revised verifier prompt and deterministic projection are evaluated
as one versioned repair package rather than as separable causal factors.

\subsection{Reference and normalized endpoints}

The \reference{} protocol installs an immutable least-privilege contract
written with the task.  It is a task-authored contract-quality control, not a
deployable baseline or a task-success ceiling: real users do not provide a
complete machine-readable contract for every edit, and overly tight contracts
can induce safe refusal.

Raw contract strings are a poor paired endpoint because semantically
equivalent constraints can differ in escaping, line context, or redundant
read-only preservation.  We therefore compare contracts after a deterministic
normalization, along two axes: \emph{protection equivalence}, whether the
contract preserves reference write isolation and required invariants, and
\emph{authorization equivalence}, whether it permits the intended minimal
edit without unnecessary restrictions.

The key mechanism test compares $C_c$ and $C_f$ within the same verified
episode.  An ideal verifier increases authorization equivalence without a
candidate-to-final protection loss.

\section{Evaluation}

\subsection{Research questions}

The primary experiment asks whether \verified{} prevents controlled concurrent
faults and terminates (RQ1), whether verification improves normalized
authorization without reducing normalized protection (RQ2), and whether its
task completion is practically non-inferior to compiler-only contracts (RQ3).
RQ2 treats the verifier as a contract-refinement mechanism; editor-accuracy
improvement is not its estimand.  A registered cross-model study then asks
whether containment and setup completeness persist under a stronger all-cell
stress claim (RQ4).  Finally, a targeted study asks whether the versioned
grounding repair closes the exposed setup boundary without a safety or
practical-utility regression (RQ5).  Each stage was frozen before execution,
following preregistration practice~\citep{vanmiltenburg2021preregistering};
the stages' distinct evidentiary roles are summarized in
Appendix~\ref{app:evidence-staging}.  We use \emph{sealed outcome} for a completed
record retained without replacement, and \emph{unsealing} for authorized access
to a previously held-out corpus.  Separately, a reviewer-requested,
post-review descriptive arm asks where genuine Git three-way merge suffices
and where instruction-conditioned authorization adds protection.

\subsection{Corpus and tasks}

The corpus uses repository-level development, validation, and test splits
drawn from licensed, compilable \LaTeX{} projects.  Method development used an
Oxford thesis template, validation used the Gemini poster template, and the
frozen test used LIPIcs and an NSFC application template.  Neither test
repository informed method development or correction.

Each repository contributes six task families: write scope, exact
preservation, citation or bibliography preservation, macro preservation,
change budget, and a cross-file consumer.  Every task specifies an intended
edit, deterministic assertions, a reference contract, and a controlled
staged fault.  Before model calls, an oracle confirms that the fault fails,
the intended edit passes, and the final project compiles using its declared
\TeX{} engine.

\subsection{Frozen matrix}

The frozen test crosses 12 tasks, two editor models
(``openai/gpt-5-mini'' and ``anthropic/claude-haiku-4.5''), three protocols,
and two repeats, for 144 episodes.  The contract compiler is
Claude Haiku 4.5 and the verifier is GPT-5 mini.  Requests use low reasoning
effort, no provider fallback, a 120-second request timeout, a 300-second
process watchdog, and an infrastructure-only retry policy.  The observed
providers were OpenAI and Amazon Bedrock through OpenRouter.

Primary outcomes are immutable first attempts.  We report task success,
committed faults, non-termination, infrastructure failure, normalized contract
endpoints, request count, and API cost.  Paired binary outcomes use exact
McNemar tests~\citep{dror2018significance}.  Task-success uncertainty uses
cluster-level bootstrap resampling over the 12 task
clusters~\citep{koehn2004significance}.

\subsection{Development control}

An earlier held-out experiment exposed two projection non-terminations.  A
subsequent pre-test validation exposed two further deterministic issues:
projection occurred after grounding, and candidate protection could retain an
exact old replacement target.  Only those six validation tasks informed this
pre-primary correction.  A fresh validation passed before source, task, fixture,
timeout, and analysis hashes were frozen for the primary test.

\subsection{Registered cross-model all-cell stress test}

After the primary systems result, we prospectively froze a stronger all-cell
stress-test claim over 30 tasks from three fresh licensed repositories.  The
study
crosses three editor models (GPT-5 mini, Claude Haiku 4.5, and Gemini 2.5
Flash), five protocol
packages (whole-file replace, unguarded exact patch, content-bound patch,
compiler-only contracts, and verified contracts), and two repeats, for 900
cells.  The compiler and verifier models, provider routing, prompts, 16-step
limit, request and process timeouts, retry policy, missingness rule,
hierarchical bootstrap seeds, ordered gates, and spend cap were frozen before
the first study completion.

The ordered claim first requires exact population and routing integrity, then
complete verified protection with zero committed faults and zero normalized
protection loss, then zero nontermination and unresolved infrastructure, and
only then tests verified-versus-replace safe task success and
candidate-to-final authorization.  No outcome is imputed; endpoints after a
failed gate remain descriptive and are not promoted to the primary claim.

\subsection{Preregistered verifier-repair study}

After sealing the all-cell outcome, we versioned the exposed grounding bug and
preregistered a targeted repair study before any model call on its selected
tasks.  Its population is all 16 exact-preservation tasks from distinct
repositories in a previously generated task pool whose local-oracle records
show task success, compilation, fault activation, repaired preview, and
stale-commit rejection.  The tasks were previously uncalled through
OpenRouter, but the corpus and oracle outcomes were already known; the study is
therefore a repair replication rather than a fresh confirmatory split.

An offline regression replays the exact overescaped verifier response that
caused the all-cell setup failure.  The live paired study asks whether the
revised prompt--projection package retains setup reliability, safety, and task
utility on previously uncalled tasks.

The study pairs \verified{} with \repair{} across GPT-5 mini, Claude
Haiku 4.5, and Gemini 2.5 Flash editors, for
$16\times3\times2=96$ episodes.  Compiler and verifier models match those of
the all-cell study, fallback is disabled, and every method-valid first attempt
is retained.  The
ordered acceptance rule requires exact population and routing, zero unresolved
infrastructure and nontermination, 48/48 complete \repair{} setup endpoints, zero
ungrounded installation and candidate-protection loss, at most one abstention,
zero committed fault, and an observed paired safe-task-success difference of
at least $-10$ points.  A 10,000-sample task-cluster bootstrap with fixed seed
is descriptive rather than a powered non-inferiority test.

\subsection{Post-review Git three-way baseline}

Following a reviewer request, we froze a separate descriptive arm before its
first baseline model completion.  It reuses the cross-model study's 30 tasks,
three editor models, and two repeats, for 180 cells.  This arm is explicitly
post-review: it is not part of the preregistered 900-cell population and does
not alter that study's ordered gates or decision.

The method creates an ephemeral Git commit of the untouched task workspace.
It records the agent's staged proposal as a binary full-index patch from that
base in a detached worktree, commits all collaborator changes as the current
pre-merge state, and makes one \texttt{git apply} attempt in
\texttt{--3way} and \texttt{--index} modes, with whitespace warnings disabled.  A clean indexed
merge commits; a conflict restores the pre-merge state without refresh or a
second attempt.  Git receives no edit contract or \system{} preview, thereby
isolating version-aware textual reconciliation from semantic authorization.
We report task success, merge status, committed controlled faults, unresolved
infrastructure, requests, and API cost.  All comparisons are descriptive.

\section{Results}

We report the registered stages before the separate post-review Git analysis,
which does not alter their decisions.  Across those stages
(Appendix~\ref{app:evidence-staging}), managed contracts always contained the
controlled faults, setup failed once and was repaired, and strict
non-inferiority remains unresolved.

\subsection{Safety, termination, and task success}

All 144 planned episodes yielded valid first-attempt outcomes.  No injected
fault committed, no episode or projection failed to terminate, and no
infrastructure failure remained unresolved.  All final workspaces compiled.

\begin{table}[t]
\centering
\caption{Frozen first-attempt test results.}
\label{tab:primary}
\small
\begin{tabular}{lrrrr}
\toprule
Protocol & Succ. & Rate & Faults & Cost (\$) \\
\midrule
\compiler{} & 46/48 & 95.83\% & 0 & 4.1292 \\
\verified{} & 46/48 & 95.83\% & 0 & 4.0162 \\
\reference{} & 44/48 & 91.67\% & 0 & 3.7013 \\
\midrule
All & 136/144 & 94.44\% & 0 & 11.8467 \\
\bottomrule
\end{tabular}
\end{table}

\verified{} succeeded on 23/24 episodes in each test repository.
\compiler{} succeeded on 22/24 LIPIcs and 24/24 NSFC episodes;
\reference{} succeeded on 21/24 and 23/24.  Notably, the task-authored
\reference{} control trails both LLM-compiled arms, illustrating the tension raised in the
introduction: a maximally tight contract can turn safety into unproductive
refusal (Section~\ref{sec:failure-analysis} details one such rollback).

\subsection{Paired contract mechanism}

\begin{table}[t]
\centering
\caption{Normalized contract endpoints within 48 verified episodes.}
\label{tab:contract}
\small
\begin{tabular}{lrrr}
\toprule
Endpoint & Cand. & Final & Losses \\
\midrule
Protection eq. & 48/48 & 48/48 & 0 \\
Authorization eq. & 24/48 & 31/48 & 0 \\
\bottomrule
\end{tabular}
\end{table}

The final contract improved authorization in seven episodes and worsened it in
none.  Conditional on the seven discordant pairs, the exact two-sided
sign/McNemar probability is $2(1/2)^7=0.015625$.  Protection equivalence
remained 48/48, so the observed improvement did not trade away the normalized
safety boundary.  The verifier modified 42/48 candidate contracts; no contract
was redeclared or relaxed while edits were staged.

\subsection{Task-success comparison}

\verified{} and \compiler{} each succeeded in 46/48 episodes, a paired
difference of 0.00 percentage points.  There were two verified-only and two
compiler-only successes (exact two-sided McNemar $p=1$).  The 12-task
cluster-bootstrap 95\% interval is $[-8.33,10.42]$ points.

The prespecified analysis plan set a non-inferiority margin of $-5$
percentage points but did not specify a confidence-interval decision rule.
The observed difference
passes the frozen operational point-estimate gate.  Under the stricter and
more conventional requirement that the confidence bound exceed the margin,
non-inferiority is not established.  We report both facts and use the latter
for confirmatory claims.  Accordingly, the supported verifier claim is
authorization refinement without observed protection loss, not a task-utility
benefit.  The descriptive stress study likewise places verified below
compiler-only task success (151/180 versus 161/180).

\subsection{Cost}

The test made 1,711 API requests and cost \$11.8467.  Verified cost decomposes
into \$3.4823 for editor trajectories, \$0.4627 for contract compilation, and
\$0.0712 for verification.  Its observed total was slightly lower than compiler-only
because stochastic editor trajectories differed; verification still has
positive intrinsic request overhead.

\subsection{Registered all-cell stress-test outcome}

The stress test is descriptively positive for managed-contract containment but
rejects the stronger claim that verified setup is complete in every cell.  Its
900-cell manifest and routing-integrity gate passed with zero deviations.  The
sealed batch contained 898 completed outcomes, one method failure, zero method
nonterminations, and one unresolved infrastructure outcome.  The observed cost
including all infrastructure retries was \$43.1748.

\begin{table}[t]
\centering
\caption{Descriptive all-cell stress-test outcomes, reported after a failed
ordered gate and not promoted to primary claims.}
\label{tab:allcell-stress}
\small
\begin{tabular}{lrrr}
\toprule
Protocol & Success & Rate & Faults \\
\midrule
Whole-file replace & 8/179 & 4.47\% & 0 \\
Unguarded exact patch & 46/180 & 25.56\% & 132 \\
Content-bound patch & 57/180 & 31.67\% & 120 \\
Compiler-only contract & 161/180 & 89.44\% & 0 \\
Verified contract & 151/180 & 83.89\% & 0 \\
\bottomrule
\end{tabular}
\end{table}

\begin{figure}[t]
\centering
\begin{tikzpicture}
\begin{axis}[
  width=0.82\linewidth, height=0.44\linewidth,
  xlabel={Committed controlled faults},
  ylabel={Task-success rate (\%)},
  xmin=-9, xmax=150, ymin=-4, ymax=102,
  grid=major, grid style={black!10},
  axis line style={black!55},
  tick style={black!45},
  ticklabel style={font=\footnotesize},
  label style={font=\small},
]
  \addplot[only marks, mark=*, mark size=2.5pt] coordinates {(0,89.44) (0,83.89)};
  \addplot[only marks, mark=square, mark size=2.4pt] coordinates {(132,25.56) (120,31.67)};
  \addplot[only marks, mark=triangle*, mark size=3pt, fill=black!45, draw=black]
    coordinates {(0,4.47)};

  \node[anchor=west, font=\scriptsize] at (axis cs:6,94) {compiler contract};
  \node[anchor=west, font=\scriptsize] at (axis cs:6,79) {verified contract};
  \node[anchor=north east, font=\scriptsize, inner sep=1pt]
    at (axis cs:143,18) {unguarded patch};
  \node[anchor=south east, font=\scriptsize, inner sep=1pt]
    at (axis cs:143,40) {content-bound patch};
  \node[anchor=west, font=\scriptsize] at (axis cs:6,9) {whole-file replace};
\end{axis}
\end{tikzpicture}
\caption{Task success versus committed controlled faults for the five
protocol packages in the descriptive 900-cell stress test.  Each package
crosses 30 tasks, three editor models, and two repeats; the whole-file
replace package has 179 resolved cells because one cell remained an
unresolved provider outcome.}
\label{fig:stress}
\end{figure}
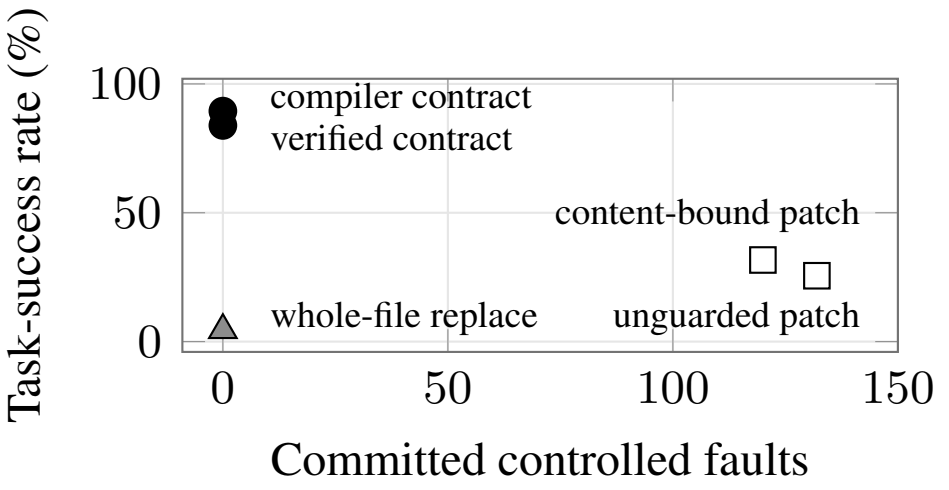

Verified contracts committed zero controlled faults and had zero observed
candidate-to-final protection losses.  Among 179 cells with complete
candidate/final normalized endpoints, authorization improved in 23 and
worsened in zero (descriptive exact two-sided sign/McNemar
$p=2.38\times10^{-7}$).  Figure~\ref{fig:stress} plots this separation: both
managed-contract packages combine high task completion with zero committed
faults, coarse patching commits more than one hundred controlled faults per
package, and whole-file replacement is safe but rarely completes the task.
However, one verified setup failed closed before a
candidate/final pair existed, so protection completeness was 179/180.  The
second ordered gate therefore failed.  A separate whole-file baseline cell
remained a provider-level infrastructure outcome after 31 total attempts.  The
preregistered stress-test decision is therefore formally inconclusive.
Recovering that baseline outcome would not change the decision because the
earlier verified setup failure already precludes the all-cell claim; the
descriptive containment and authorization results remain as narrower evidence.

\subsection{Verifier-repair closure}

The repair study tests this setup-boundary closure without a safety or
practical-utility regression; it does not reopen the
sealed stress-test decision.  All 96 cells completed with exact population and
routing.  There were no infrastructure failures or method nonterminations,
and the observed cost was \$3.9411.  Table~\ref{tab:repair} reports the frozen
protocol endpoints.

\begin{table}[t]
\centering
\caption{Preregistered verifier-repair outcomes.}
\label{tab:repair}
\small
\setlength{\tabcolsep}{3pt}
\begin{tabular}{@{}>{\raggedright\arraybackslash}p{0.56\linewidth}rr@{}}
\toprule
Endpoint & \verified{} & \repair{} \\
\midrule
Episodes & 48 & 48 \\
Valid installed contracts & 48 & 48 \\
Explicit abstentions & 0 & 0 \\
Ungrounded installed preservation & 0 & 0 \\
Candidate-protection losses & 0 & 0 \\
Committed controlled faults & 0 & 0 \\
Safe task successes & 48 & 47 \\
Mean API cost (USD) & 0.04214 & 0.04093 \\
\bottomrule
\end{tabular}
\end{table}

All six ordered gates passed, so the versioned repair met its preregistered
acceptance criteria.  Every \repair{} cell installed a valid grounded
contract, so no abstention was needed.  The paired safe-task-success difference
was $-2.08$ points and the descriptive one-sided 95\% task-cluster bootstrap
lower bound was $-6.25$ points; the observed difference passes the frozen
$-10$ point practical guardrail.

No live \repair{} response contained an ungrounded verifier-only atom, so the
grounding projection recorded zero natural corrections.  The exact all-cell
overescaping response is instead covered by an offline regression test: \repair{}
drops the ungrounded verifier atom, restores the compiler's grounded atom, and
installs a valid contract in one attempt.  The live study therefore
establishes repair compatibility on previously uncalled tasks; it does not
estimate the natural frequency of the rare encoding failure or provide new
population-wide efficacy evidence.

\subsection{Post-review Git baseline and component comparison}
\label{sec:git-baseline}

All 180 Git cells produced method-valid outcomes, with no unresolved
infrastructure or nontermination.  The arm made 1,198 requests and cost
\$4.7090.  Git achieved 53/180 task successes (29.44\%) and committed 123
controlled faults.  The agent reached the merge step in 176 episodes; every
one produced a clean merge, while four episodes never attempted commit.  Thus,
a clean textual merge was not an authorization or preservation proof.

The family breakdown identifies where Git is the appropriate mechanism.  It
succeeded on all 36/36 cells in post-preview version drift and same-file
non-overlapping concurrency, committing no controlled fault there.  Across the
remaining semantic-authorization and preservation families, it succeeded in
17/144 cells and committed 123 controlled faults.  In particular, success was
0/18 in each of changed-line budget, cross-file consumer, and write-set scope.
Its aggregate success was comparable to unguarded exact patching (46/180) and
content-bound patching (57/180), which committed 132 and 120 faults,
respectively.

Table~\ref{tab:component-comparison} separates the component comparison into
population-matched panels.  Panel~A shows that version awareness alone handles
textual reconciliation but not semantic write authority.  Panel~B shows that
the reference is not an oracle performance ceiling: a task-authored contract
can be sufficiently tight to induce safe but unproductive refusal.  Within the
48 verified episodes, independent verification improved normalized
authorization in 7/48 contracts, caused zero candidate-to-final protection
losses, and left task success equal at 46/48.

\begin{table}[t]
\centering
\caption{Population-matched component comparisons.  Panel A includes the
post-review descriptive Git arm; Panel B reports the primary contract-quality
control.  This analysis was assembled after review from the Git arm and
the original registered contract outcomes; it compares protocol packages
rather than isolating causal effects of individual components.}
\label{tab:component-comparison}
\small
\setlength{\tabcolsep}{3pt}
\begin{tabular}{@{}>{\raggedright\arraybackslash}p{0.59\linewidth}rr@{}}
\toprule
Component setting & Success & Faults \\
\midrule
\multicolumn{3}{@{}l}{\emph{Panel A: 180-cell cross-model population}} \\
Version-aware Git merge without semantic contract & 53/180 & 123 \\
Compiled contract without verifier & 161/180 & 0 \\
Compiled and independently verified contract & 151/180 & 0 \\
\midrule
\multicolumn{3}{@{}l}{\emph{Panel B: 48-cell primary population}} \\
Compiled contract without verifier & 46/48 & 0 \\
Compiled and independently verified contract & 46/48 & 0 \\
Task-authored reference contract & 44/48 & 0 \\
\bottomrule
\end{tabular}
\end{table}

\section{Failure Analysis}
\label{sec:failure-analysis}

In the 144-episode primary test, eight episodes missed at least one
requested-edit assertion while committing no injected fault.  Three GPT-5
mini trajectories wrote \verb|\\cite| instead
of \verb|\cite|.  Compilation accepted the resulting source, illustrating why
artifact compilation alone is insufficient.  Two Claude Haiku trajectories
reached the step limit without commit.  One reference-contract trajectory
detected a concurrent macro fault and rolled back its staged edit: safe, but
not task-complete.

The remaining two failures reveal a compiler-only method boundary.  The
instruction authorized replacement of \verb|\title{Dummy title}|, while the
compiler preserved the longer source fragment
\verb|\title{Dummy title} %TODO Please add|.  The exact old-side projector did
not recognize this suffix-bearing atom as edit-conflicting.  One editor
refused; another retained the old title in a comment and failed an assertion.
The verified arm succeeded on both corresponding cells.

The frozen automated taxonomy counts these completed episodes as task
failures, not setup-level method failures; we retain that preregistered label
and disclose the semantic projection limitation.  A suffix-aware compiler
repair would be a separately versioned method and must be evaluated on fresh
tasks.

The registered all-cell stress test exposed a second boundary in probabilistic
contract setup.  In one verified cell, deterministic verification rejected an
ungrounded preservation fragment after both allowed verifier attempts.  The
system failed closed before the editor transaction and committed no fault, but
the candidate/final normalized contract endpoints were consequently absent.
This is preferable to installing an ungrounded contract, yet it violates the
strong study's all-cell mechanism-completeness requirement.  We retain it as a
method failure rather than retrying it away.

The \repair{} study had one task failure but no setup or safety failure.  In
one OpenAI-editor episode, contract setup was valid,
grounded, and protection-complete.  After preview exposed the injected removal
of a protected subsection, the editor repaired that invariant but then used
the unqualified phrase \verb|What is Deep Learning?| with occurrence one.  It
changed the same phrase in a leading comment rather than the still-unchanged
\verb|\section{...}| command.  Preview correctly accepted the grounded,
contract-compliant two-line edit; final task assertions rejected the missed
requested heading.  The paired \verified{} cell succeeded.  We retain the outcome
as a task failure and attribute it to target disambiguation during editor fault
recovery, not to verifier grounding.

The sole unresolved stress-test cell used Gemini 2.5 Flash with whole-file
replace.  Thirty-one attempts ended with the provider-level
\verb|finish_reason=error| at varying trajectory steps.  We neither impute an
outcome nor relabel it as a method failure.  Further retries were stopped once
the earlier ordered-gate failure made them decision-irrelevant.

Attempts to construct a larger fresh confirmatory population did not pass the
frozen feasibility gates, and we did not adapt or retry those outcomes in
place.  Strict non-inferiority therefore remains open for lack of an admissible
powered population; Appendix~\ref{app:feasibility} records the screening
chronology.

\section{Related Work}

\paragraph{Agent benchmarks.}
SWE-bench grounds agent evaluation in repository-level edits and executable
tests~\citep{jimenez2023swebench}.  ToolSandbox tests state dependencies and
intermediate milestones~\citep{lu2024toolsandbox}; $\tau$-bench tests domain
policies and database goal states~\citep{yao2024taubench}.  \system{} is narrower in domain but adds
version-bound write authorization, controlled concurrency faults, and paired
candidate-to-final contract endpoints.

\paragraph{Delegation and diagnostics.}
DELEGATE-52 measures semantic document corruption across repeated delegated
transformations~\citep{laban2026delegate}.  Token-probability diagnostics
characterize instability within reasoning
traces~\citep{chen2026instability}.  \system{} complements these behavioral
analyses by checking whether an external state transition is authorized under
concurrent workspace changes.

\paragraph{Contracts and agent security.}
Edit contracts adapt design-by-contract
interfaces~\citep{meyer1992contract} to probabilistic editors.
ContractBench studies observation contracts such as expiring or byte-sensitive
API artifacts~\citep{wang2026contractbench}.  Its contracts constrain how an
agent reuses tool observations.  \system{} studies edit contracts that
authorize a persistent multi-file state transition under concurrency.
Indirect prompt injection compromises LLM-integrated
applications~\citep{greshake2023injection}.  AgentDojo benchmarks attacks and
defenses for tool-using agents~\citep{debenedetti2024agentdojo}; ToolEmu surfaces
risks in an emulated sandbox~\citep{ruan2024toolemu}.  CaMeL
separates control and data flow and uses capabilities to constrain
unauthorized information flow~\citep{debenedetti2025camel}; Progent and
emerging permission models impose privilege control on agent
actions~\citep{shi2025progent,wu2026agentperms}.  Unlike general capability
systems that govern tool calls or information flow, \system{} applies
established least privilege~\citep{saltzer1975protection} to one concrete
persistent mutation; it is not a new general permission model.

\paragraph{Collaborative systems and version control.}
Optimistic concurrency control validates work before commit rather than
locking all data throughout a transaction~\citep{kung1981optimistic}.
Operational transformation and conflict-free replicated data types address
convergence and intention preservation for concurrent edits
\citep{ellis1989ot,sun1998achieving,shapiro2011crdt,preguica2018crdt}.
Git three-way merge tests textual compatibility: our post-review arm succeeded
on all 36 concurrency cells yet committed 123 faults elsewhere despite clean
merges.  \system{} complements Git, operational transformation, and replicated
data types by adding instruction-conditioned authorization above their merge
and convergence mechanisms.

\section{Conclusion}

\system{} treats an LLM document edit as a transaction rather than a trusted
text response.  In the primary test, its version-bound contract, immutable
staged state, deterministic preview, and exact commit boundary contained every
controlled fault.  Semantic verification acted as a contract-refinement
mechanism, improving authorization in seven paired episodes without measured
protection loss; equal 46/48 task success did not establish a utility benefit.

The cross-model stress test widened the evidence with unmanaged baselines:
contract protocols committed zero controlled faults while coarse patching
committed more than one hundred per package.  One fail-closed method failure
and one unresolved provider outcome left the stronger all-cell claim
inconclusive.  A separately versioned repair achieved 48/48 grounded setups
without protection loss, fault, or abstention; it did not reopen the sealed
decision.

The post-review Git arm sharpened the systems boundary.  Three-way merge
completed all 36 textual-concurrency cells, but 176 clean merges across the
full matrix still committed 123 controlled faults.  Version control therefore
supplies textual merge mechanics; the contract layer supplies the measured
authorization and preservation boundary.

The qualified systems claim is containment of these benchmark faults; setup
and task completion remain model-dependent, and strict non-inferiority needs a
fresh, powered population.  Evidence is limited to controlled faults in
localized \LaTeX{} edits: compilation, exact preservation, and deterministic
assertions do not establish semantic or factual correctness or user
satisfaction.  The open frontier is compiling permissive yet safe contracts
and testing longer-horizon workflows.

\section{Limitations}

The primary test has only 12 independent task clusters and two held-out
repositories.  Repeats and models increase episode count but do not create
independent task diversity, so the task-success interval remains wide.  Our
attempts to assemble a larger admissible corpus did not satisfy the frozen
feasibility gates.  Future non-inferiority work should power on repository and
task clusters and preregister a one-sided confidence-bound rule.

All tasks are localized \LaTeX{} edits with deterministic assertions.
Compilation and exact preservation do not measure prose quality, factual
correctness, or user satisfaction.  Exact assertions catch byte-level errors
that compilation misses, but future work should add semantic and human-reviewed
outcomes without replacing deterministic safety endpoints.
Our fixed-task evaluation also leaves open-ended agent--environment
co-evolution untested, a broader challenge identified for multi-agent
simulations~\citep{chen2025staticsandboxes}.

The evaluated models cover only part of the API ecosystem, and behavior of
API-served models can change over time~\citep{chen2024chatgpt}.  Persisted response identifiers, observed providers, trajectories,
costs, and configuration hashes improve reproducibility but cannot reconstruct
proprietary weights.  One provider outcome also remained unresolved in the
all-cell study.

The repair study uses one exact-preservation task per repository and one repeat
per editor--protocol cell.  Its tasks were uncalled by models but selected from
a previously unsealed corpus with known local-oracle outcomes, and the rare
overescaping error did not recur naturally.  The study supports repair totality
and compatibility, not a natural failure rate or population-wide ranking.

Finally, reference contracts are task-author artifacts and may encode author
bias.  Normalized equivalence depends on those references and on our chosen
write-isolation abstraction.  ``Zero faults'' means zero controlled benchmark
faults under these tasks, not proof of universal safety.

\section{Ethical Considerations}

The corpus uses licensed public templates, records upstream provenance, and
excludes repositories requiring unavailable proprietary assets.  It contains
no personal or confidential documents.  Model calls nevertheless transmit
task content to API providers, so the current system should not be applied to
sensitive workspaces without an appropriate data-processing agreement.  More
generally, the transaction layer limits specified writes but cannot guarantee
the factual quality or downstream appropriateness of an authorized edit.

\bibliography{references}

\appendix

\section{Evidence Staging and Claim Boundaries}
\label{app:evidence-staging}

Table~\ref{tab:appendix-evidence-map} distinguishes the three registered studies
from the separate post-review Git arm and retains each study's claim boundary.
The registered stages are the frozen
\verified{} primary test, the registered cross-model all-cell stress test, and
the preregistered \repair{} verifier-grounding repair study.  The post-review
Git arm provides descriptive version-control evidence outside the registered
gate sequence.

\begin{table*}[!b]
\centering
\caption{Evidence progression and the claim assigned to each study.}
\label{tab:appendix-evidence-map}
\small
\begin{tabular}{@{}>{\raggedright\arraybackslash}p{0.15\textwidth}
                    >{\raggedright\arraybackslash}p{0.28\textwidth}
                    >{\raggedright\arraybackslash}p{0.49\textwidth}@{}}
\toprule
Stage & Purpose & Frozen evidentiary status \\
\midrule
Primary (144 episodes) & Core transaction and verifier mechanism &
Containment and paired authorization improvement supported; strict
task-success non-inferiority unresolved. \\
All-cell stress (900 cells) & Cross-model and baseline robustness &
Descriptively positive containment and authorization results; stronger
all-cell claim formally inconclusive after incomplete verified endpoints and
one unresolved provider outcome. \\
Repair compatibility (96 episodes) & Versioned grounding repair and regression
risk & Repair acceptance criteria satisfied for setup totality, safety, and the prespecified
practical-utility guardrail; not population-wide efficacy evidence. \\
Post-review Git (180 episodes) & Reviewer-requested version-control baseline &
Descriptive evidence only: textual concurrency succeeded in 36/36 cells, but
clean merges admitted 123 controlled faults; outside the preregistered gate
sequence. \\
\bottomrule
\end{tabular}
\end{table*}

\section{Frozen Corpus-Expansion Chronology}
\label{app:feasibility}

The larger frozen efficacy test was attempted but did not survive the
prespecified feasibility gates.  One sealed generation attempt halted when a
confirmatory repository could not instantiate a required task family.  A
subsequent screen of 240 fresh exact-commit candidate repositories admitted 17
of the required 28.  The expanded task pool
later reached 28 repositories, but a storage-capacity failure during
confirmatory oracle validation rendered that run inadmissible under the
frozen deviation policy.  A final replacement
screen of 186 further candidates yielded three eligible repositories.  Under
the frozen deviation policies, none of these outcomes was adapted or retried
in place.  Corpus construction is therefore retained as a negative feasibility
result under the admission gates: strict non-inferiority remains open for
structural reasons, not for lack of attempting a larger test.

\section{Artifact Contents and Availability}
\label{app:artifact}

We have prepared a reproducibility package containing source code, schemas, licensed
fixtures, task generators, frozen configurations, persisted first-attempt
analyses, candidate and final contract endpoints, and an API-only experiment
runner.  It also includes the frozen Git-arm configuration, implementation,
persisted outcomes, and population-matched comparison analysis.  At the time
of writing, these materials have not been publicly released; no public
repository or DOI is available.  Offline reproduction from the persisted
records requires no API access.  Generating
fresh trajectories requires an OpenRouter key but no local model serving or
HPC resources.

\end{document}